\documentclass[journal]{IEEEtran}

\ifCLASSINFOpdf
\else
\fi

\usepackage{cite}
\usepackage{times}
\usepackage{algorithmic}
\usepackage{array}
\usepackage{amssymb}
\usepackage{mdwmath}
\usepackage{mdwtab}
\usepackage{eqparbox}
\usepackage{cite}
\usepackage{url}
\usepackage{multicol,multirow}
\usepackage[cmex10]{amsmath}
\usepackage{makeidx}
\usepackage{diagbox}
\usepackage{rotating,soul}
\usepackage{wrapfig}
\usepackage{booktabs}
\usepackage[usenames,dvipsnames]{color}
\usepackage{ragged2e}

\usepackage{subcaption}
\usepackage{tabularx}
\usepackage[normalem]{ulem}
\usepackage{booktabs}
\usepackage{multirow}
\usepackage{algorithm}
\usepackage{amsmath}

\usepackage{pifont}
\usepackage{overpic}
\makeindex

\usepackage{geometry}
\RequirePackage{silence}
\DeclareGraphicsExtensions{.pdf,.jpg,.png}

\usepackage[colorlinks,bookmarksnumbered,bookmarksopen,linkcolor=red,citecolor=blue,urlcolor=blue]{hyperref}

\usepackage{pgfplots}
\pgfdeclarelayer{background}
\pgfdeclarelayer{foreground}
\pgfsetlayers{background,main,foreground}

\definecolor{mygray}{gray}{.92}

\def\etal{{\em et al.}}

\newcommand{\addFig}[1]{}
\newcommand{\addFigs}[1]{}

\graphicspath{{./figs/}}

\newlength\savedwidth

\begin{document}

\title{Enhancing Medical Visual Grounding via Knowledge-guided Spatial Prompts}

\author{
Yifan Gao, Tao Zhou, Yi Zhou, Ke Zou, Yizhe Zhang, Huazhu Fu
\thanks{Y. Gao, T. Zhou, and Y. Zhang are with the School of
Computer Science and Engineering, Nanjing University of Science and Technology, Nanjing 210094, China. Y. Zhou is with the School of Computer Science and Engineering, Southeast University, Nanjing 211189, China. K. Zou is with Yong Yoo Lin School of Medicine, National University of Singapore, Singapore. H. Fu is with the Institute of High-Performance Computing, Agency for Science, Technology and Research, Singapore. Corresponding authors: Tao Zhou (taozhou.dreams@gmail.com).}
}

\markboth{}%
{**}

\maketitle

\begin{abstract}
Medical Visual Grounding (MVG) aims to identify diagnostically relevant phrases from free-text radiology reports and localize their corresponding regions in medical images, providing interpretable visual evidence to support clinical decision-making. Although recent Vision–Language Models (VLMs) exhibit promising multimodal reasoning ability, their grounding remains insufficient spatial precision, largely due to a lack of explicit localization priors when relying solely on latent embeddings. In this work, we analyze this limitation from an attention perspective and propose \textbf{KnowMVG}, a Knowledge-prior and global–local attention enhancement framework for MVG in VLMs that explicitly strengthens spatial awareness during decoding. Specifically, we present a knowledge-enhanced prompting strategy that encodes phrase related medical knowledge into compact embeddings, together with a global–local attention that jointly leverages coarse global information and refined local cues to guide precise region localization. This design bridges high-level semantic understanding and fine-grained visual perception without introducing extra textual reasoning overhead. Extensive experiments on four MVG benchmarks demonstrate that our KnowMVG consistently outperforms existing approaches, achieving gains of 3.0\% in AP50 and 2.6\% in mIoU over prior state-of-the-art methods. Qualitative and ablation studies further validate the effectiveness of each component. The code will be released upon acceptance at \href{https://github.com/taozh2017/KnowMVG}{https://github.com/taozh2017/KnowMVG}.

\end{abstract}

\begin{IEEEkeywords}
Federated semi-supervised learning, global-local interaction, medical image segmentation
\end{IEEEkeywords}

\IEEEpeerreviewmaketitle

\section{Introduction}\label{introduction}

\begin{figure}
	\centering
	\includegraphics[width=\columnwidth]{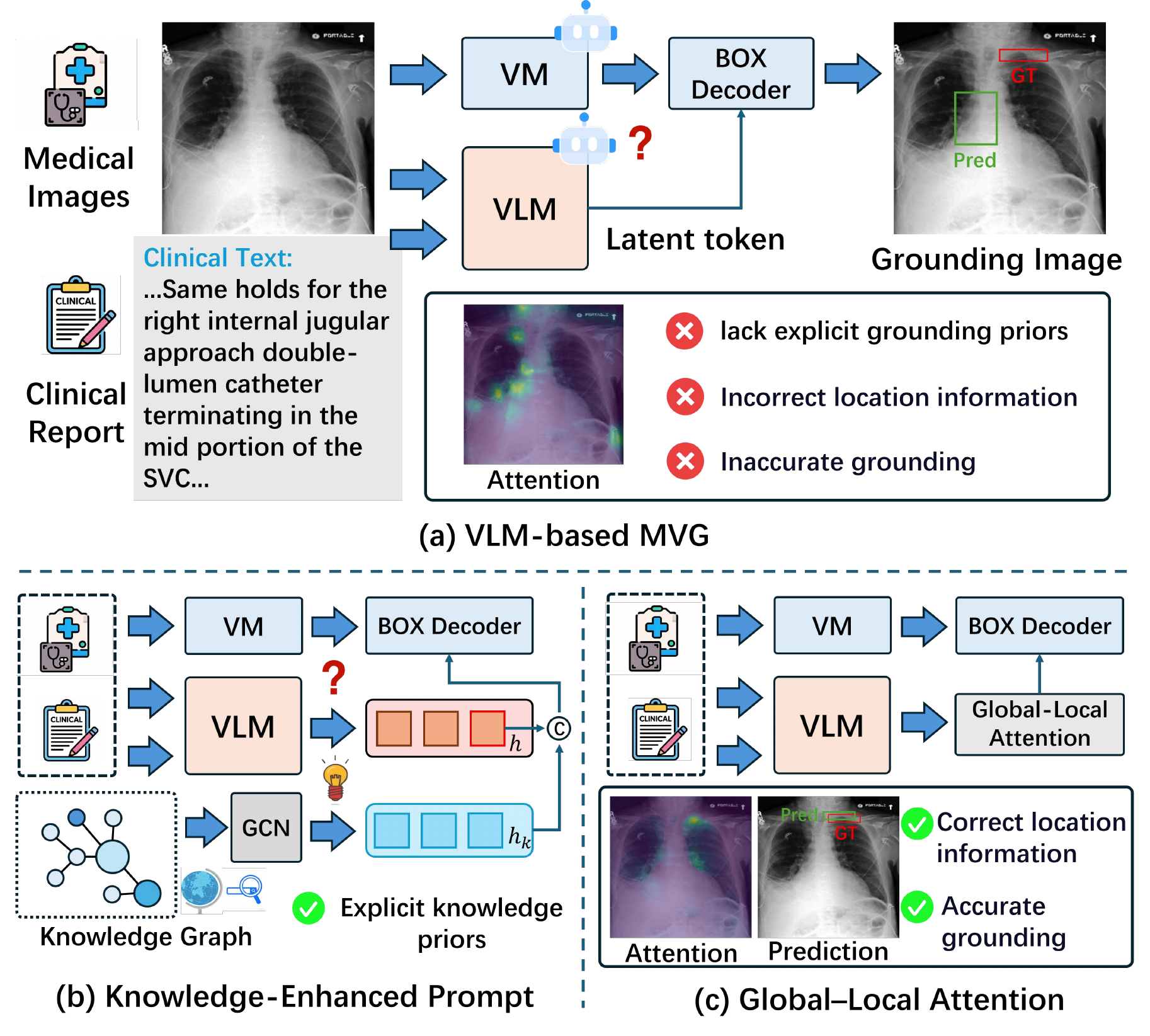} 
    \caption{Motivation and architectural comparison for medical visual grounding. (a) Existing VLM-based MVG relies on latent token prompts with unstable attention. (b) Our knowledge-enhanced prompting strategy injects phrase-level priors for improved grounding ability. (c) Global-local attention module combines semantic masks and box cues to achieve accurate and consistent grounding.}
	\label{intro-1}
\end{figure}

Medical Visual Grounding (MVG) aims to localize clinically relevant regions in medical images based on semantic descriptions and has become an important tool for assisting radiological interpretation. Early studies focused on phrase-level grounding \cite{chen2023medical}, where diagnostic phrases are extracted from reports and localized through multi-stage pipelines. While effective in controlled settings, such approaches rely on explicit phrase annotations and suffer from limited scalability and clinical practicality. To overcome these limitations, recent work has advanced toward MVG \cite{zou2025uncertainty}, which directly associates free-text reports with visual evidence in a unified framework. As illustrated in Fig.~\ref{intro-1}~(a), MVG jointly identifies diagnostically relevant textual units and grounds them spatially, better reflecting real-world clinical workflows. 

\label{sec:introduction}

Although visual grounding has been extensively studied in the natural image domain \cite{deng2021transvg,zhu2022seqtr,he2024improved,xiao2024hivg,kang2024segvg}, medical grounding remains considerably more difficult due to limited annotations, subtle pathological cues, and strong anatomical priors. Early medical approaches \cite{chen2023medical,ichinose2023visual} often relied on handcrafted preprocessing or rule-based phrase extraction, limiting robustness under diverse report styles. The emergence of large Vision-Language Models (VLMs) \cite{achiam2023gpt,liu2023visual,chen2024internvl,chen2024far} has enabled report-level multimodal reasoning and simplified grounding pipelines via latent token representations. However, as illustrated in Fig.~\ref{intro-1}~(a), despite improved diagnostic recognition, their predicted bounding boxes frequently misalign with true pathological regions, highlighting a critical gap between semantic understanding and precise spatial localization.

To better understand this issue, we analyze the attention behavior of the box decoder in Fig.~\ref{intro-1}. The results show that latent token prompts fail to explicitly encode object-level spatial information, forcing the decoder to rely primarily on inherent inductive biases and subsequent refinement. Further analysis in Fig.~\ref{intro-1}~(c) reveals that introducing explicit spatial cues lead to significantly more focused and semantically aligned attention. These observations highlight a fundamental limitation of purely latent token prompting mechanisms: \textbf{they lack the capacity to convey explicit grounding information, which is essential for precise localization in medical images}.

Motivated by these findings, we argue that effective MRG requires explicit and structured grounding cues that bridge high-level semantic understanding and low-level spatial localization. To this end, we propose a Knowledge-enhanced prompting strategy and global–local attention enhancement framework for Medical Visual Grounding (KnowMVG), which fully leverages knowledge priors and multi-scale attention to enhance spatial awareness during decoding. 

Specifically, we present a Knowledge-enhanced Prompting Strategy (KPS) to selectively extract grounding-related information tied to key diagnostic phrases, forming structured knowledge priors. Unlike conventional Retrieval Augmented Generation (RAG), we do not directly feed these grounding priors into the VLM as long textual inputs. Instead, we encode these priors into compact embeddings via a lightweight graph convolution operation, which are then used as knowledge-enhanced prompts, thereby avoiding the reasoning overhead associated with long contextual inputs. Furthermore, based on our analysis, latent token prompts can misdirect the decoder's attention. In contrast, explicit visual cues serve as local attention prompts, offering coarse but intuitive grounding information about object locations. This mechanism aligns with clinical reasoning patterns, such as the intuitive judgment that ``pneumonia occurs in the lung, specifically in the left lower lobe”. Building on this, we propose a Global–Local Attention (GLA) module. It leverages refined global visual cues to constrain the target's plausible region, while local localization provide explicit grounding cues. This hybrid attention guidance enables the decoder to first narrow candidate regions and then refine object grounding, significantly enhancing its capability.

Our contributions can be summarized as follows:

\begin{itemize}
\item We systematically analyze the limitations of visual grounding in MVG and identify a critical attention deficiency in the decoder stage that hampers accurate grounding. This finding provides a new perspective for understanding the underlying causes of grounding errors and opens up new avenues for improving grounding performance.
\item We propose a knowledge-enhanced prompting strategy that explicitly injects phrase-related medical knowledge into the grounding process, providing prior-guided object constraints and reliable localization cues beyond latent embeddings.
\item We present a global-local attention module that synergistically integrates global context with fine-grained local spatial cues, enhancing feature discrimination and enabling more precise and robust visual grounding.
\item Extensive experiments on four MVG benchmarks, including MRG-MS-CXR, MRG-CHESTX-RAY8, MRG-MIMIC-VQA, and MRG-MIMIC-CLASS, demonstrate that our method achieves superior performance in enhancing the grounding ability of VLMs. 
\end{itemize}

\vspace{5pt}
\section{Related Work} \label{related work}
\subsection{Vision-Language Models}
\vspace{5pt}

Recent large-scale Vision--Language Models (VLMs) 
\cite{achiam2023gpt,liu2023visual,chen2024internvl,chen2024far,sun2023emu,sun2024generative,wang2024qwen2,team2023gemini,lin2024vila,comanici2025gemini}, have significantly advanced multimodal understanding and reasoning by leveraging large language models as powerful semantic backbones. These models demonstrate impressive capabilities in cross-modal reasoning, instruction following, and open-ended visual question answering. The training alignment process of the GPT-4 \cite{achiam2023gpt} family is designed to improve the model’s adherence to desired behaviors. Within the open-source ecosystem, the LLaVA \cite{liu2023visual} series introduces visual instruction tuning, enabling stronger multi-turn dialogue and more general vision–language understanding capabilities. InternVL \cite{chen2024internvl} approaches this problem from the perspective of matching parameter scale with representational capacity by scaling up the visual encoder and adopting a progressive image–text alignment strategy to narrow the capability gap between the visual encoder and the LLM. In addition, Gemini \cite{team2023gemini} proposes a family of natively multimodal models, emphasizing end-to-end multimodal modeling to achieve stronger cross-modal understanding and reasoning abilities. To bridge visual encoders and language models, LISA \cite{lai2024lisa} introduces latent embedding prompts, enabling more flexible interactions between visual features and language representations. Despite these advances, most generic VLMs are optimized for semantic reasoning rather than precise spatial localization. Their reliance on implicit attention over latent tokens often results in diffuse or unstable grounding, especially in the field of medical imaging where subtle visual cues demand high spatial precision.

\subsection{Medical Visual Grounding}
\vspace{5pt}

Visual grounding aims to localize entities specified by natural language expressions and has been extensively studied in the natural image domain. Existing methods have evolved from early two-stage pipelines to unified multimodal Transformer architectures \cite{wang2019phrase,datta2019align2ground,zhou2023joint,gu2024context}, learning joint vision–language representations that achieve strong benchmark performance. In contrast, medical visual grounding remains underexplored due to scarce annotations and fragmented supervision. Available datasets often provide rich clinical narratives but weak spatial alignment, hindering models from learning clinically meaningful cues such as morphology, laterality, and anatomical sublocations \cite{li2022grounded,kamath2021mdetr,liu2023qilin,yang2025new}.

Early medical approaches relied on structured multi-stage pipelines \cite{pellegrini2023radialog,chen2023medical}. For example, MedRPG \cite{chen2023medical} extracts diagnostic phrases before regressing bounding boxes, while such designs heavily depend on report preprocessing and handcrafted components. More recent studies leverage large language models (LLMs) to process free-text reports~\cite{deng2025med,zhang2025anatomical,zhang2026medground,vilouras2024zero,he2025parameter,xu2025medground,wang2025endochat,jing2025reason}. MedGround \cite{zhang2026medground} explicitly train the model to follow clinically grounded referring descriptions by converting segmentation annotations into scalable image–text–box supervision. Zou \etal~\cite{zou2025uncertainty} leverage LLMs to extract diagnostically meaningful phrases and introduce uncertainty-aware prediction to enhance robustness, whereas Vilouras \etal~\cite{vilouras2024zero} investigate zero-shot medical phrase grounding using generative foundation models. Nevertheless, existing methods often adapt generic VLMs through heuristic modifications without explicitly incorporating medical knowledge priors, leading to suboptimal localization accuracy and limited interpretability.

\begin{figure*}[t]
	\centering
	\includegraphics[width=1.0\textwidth]{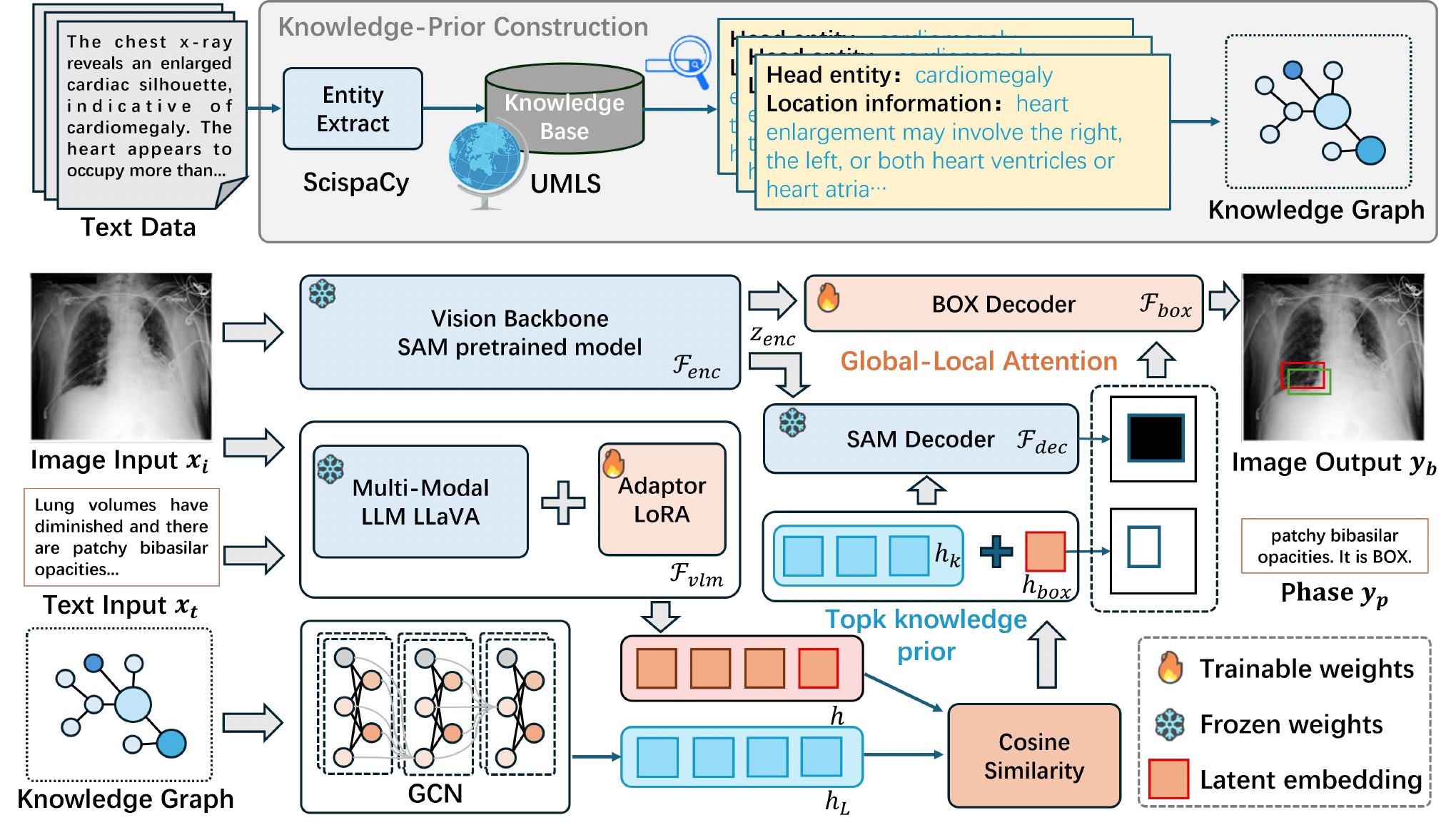} 
	\caption{Overall framework of the proposed medical report grounding method. Given a medical image and its corresponding clinical text, a vision backbone and a multimodal LLM jointly encode visual and textual representations. To enhance spatial grounding, we introduce a knowledge-enhanced prompting strategy that encodes phrase-related anatomical information and selects top-k location prompts. These priors are combined with a global–local attention module to guide the BOX decoder toward clinically relevant regions.}
	\label{overview}
\end{figure*}

\section{Method}
\subsection{Preliminary}
\vspace{5pt}

Medical visual grounding aims to extract key diagnostic information from an X-ray image $x_i$ and its corresponding raw medical reports $x_t$ in a manner that emulates radiologists’ expert interpretation. The objective is to identify medical phrases $y_p$ mentioned in the report and to localize their corresponding regions in the X-ray image via bounding boxes $y_b$. This process can be expressed as follows:
\begin{equation}
    ({y_p},{y_b}) = \mathcal{F}({x_i},{x_t}),
\end{equation}
where $\mathcal{F}(\cdot)$ represents the mapping function that relates the input variables to the output. 

Based on our prior attention analysis, we identify that this limitation primarily stems from relying solely on latent token embeddings as guidance for the box decoder, which are poorly aligned with the pretrained priors of the segmentation-based decoder. Such embeddings fail to provide sufficiently informative spatial cues to the visual representations. Motivated by these findings, we design two key prompts, Knowledge-enhanced Prompting Strategy (KPS) and Global–Local Attention (GLA) to enhance the grounding capability of VLMs. The implementation details of these components are presented in Sections C and D, respectively.

\subsection{Overview of Our Method}
\vspace{5pt}

As illustrated in Fig.~\ref{overview}, to strengthen the model’s ability to capture localization cues during training, we introduce top-$k$ knowledge-priors $e_k$ together with coarse visual grounding derived from latent token embeddings. Prior work \cite{zou2025uncertainty} has demonstrated that latent token embeddings can encode local visual information. We further incorporate these cues as prompts to guide the box decoder toward relevant regions. Meanwhile, the SAM-pretrained decoder utilizes prompt embeddings to suppress background noise and generate more focused spatial responses. To avoid disrupting its pretrained priors, we adopt a dual-decoder architecture. Specifically, the SAM decoder first produces a global visual prompt with reduced background interference, which is then combined with local prompts and fed into the box decoder. This global–local design enables more accurate and robust grounding.

\subsection{Knowledge-prior Construction}
\vspace{5pt}

Our primary objective is to enhance the grounding capability of VLMs by learning latent embedding representations that explicitly encode localization priors. To this end, we construct a localization-oriented knowledge graph that captures phrase-specific prior knowledge for each individual dataset. 

\subsubsection{Entity Extraction}
\vspace{5pt}

In our study, we conduct experiments on four datasets: MRG-MS-CXR, MRG-ChestX-ray8, MRG-MIMIC-VQA, and MRG-MIMIC-Class, which encompass images, reports, questions, and answers related to radiology and pathology. To collect key entities from these datasets, we adopt a structured approach to extract domain-specific information from the training split. As illustrated in Fig.~\ref{overview}, given a dataset $\mathcal{D} = \{(Q_i, A_i)\}_{i=1}^{M}$, where $Q_i$ and $A_i$ denote the report and answer, respectively. We extract domain-specific entities using SciSpaCy \cite{neumann-etal-2019-scispacy}, a biomedical entity parser can be obtained by
\begin{equation}
\mathcal{E}_i = \text{SciSpaCy}(Q_i) \cup \text{SciSpaCy}(A_i). 
\end{equation}

Earlier studies have shown that more refined textual content can reduce information redundancy and provide more precise guidance \cite{huang2024enhancing}. Therefore, for each extracted entity $e \in \mathcal{E}$, where $\mathcal{E} = \bigcup_{i=1}^{M} \mathcal{E}_i$, we query the Unified Medical Language System (UMLS) \cite{bodenreider2004unified} to retrieve its localization-related descriptions:
\begin{equation}
L_e = \text{UMLS}(e),
\end{equation}
where $L_e$ denotes the set of localization-related information linked to entity $e$. 

\subsubsection{Knowledge Graph}\vspace{5pt}
Based on the retrieved information, the knowledge graph is constructed as a set of triples by
\begin{equation}
KG = \{ (e_h, r, l) \mid e_h \in \mathcal{E},\ l \in L_{eh} \},
\end{equation}
where $e_h$ denotes the head entity, $l$ represents the corresponding localization-related information, and $r$ indicates the predefined relation linking entities to their localization descriptions. The constructed knowledge graph is further represented as a graph $\mathcal{G} = (\mathcal{V}, \mathcal{E})$, where each node corresponds to an entity or its localization information. The relational structure is encoded by an adjacency matrix, which can be expressed by
\begin{equation}
X_{adj}(i,j) =
\begin{cases}
1, & \text{if a relation exists between node } i \text{ and } j, \\
0, & \text{otherwise}.
\end{cases}
\end{equation}

\vspace{5pt}
\subsection{Grounding Enhancement}
\vspace{5pt}

\subsubsection{Knowledge-enhanced Prompting Strategy}
\vspace{5pt}
In the previous subsection, we have constructed a knowledge graph for each dataset, consisting of key entities and their corresponding localization-related information. Unlike conventional RAG approaches, we do not incorporate this knowledge as additional textual input to the VLM, as textual injection neither guarantees spatially grounded representations nor aligns with the localization objective. Such a strategy would also increase inference overhead. To effectively exploit the constructed knowledge graphs, we encoded each node in the knowledge graph into a partial embedding using RoBERTa \cite{liu2019roberta} by
\begin{equation}
\mathbf{h}_i^{(0)} = \mathcal{F}_{\text{roberta}}(t_i),
\end{equation}
where $t_i$ denotes the textual content of the $i$-th node in $KG$. Given the adjacency matrix $X_{adj}$, a graph convolutional network $GCN$ is applied to propagate localization information across the graph by
\begin{equation}
\mathbf{h}_L = GCN(\{\mathbf{h}_i^{(0)}\}, X_{adj}).
\end{equation}

However, directly using all latent representations $\mathbf{h}_L$ as prompts would inevitably introduce unnecessary noise. To address this issue, we compute the cosine similarity between each localization-aware embedding $\mathbf{h}_i^{(L)} \in \mathbf{h}_L$ and the multimodal output embedding $\mathbf{h}_{vlm}$ as follows
\begin{equation}
s_i = \frac{\mathbf{h}_i^{(L)\top} \mathbf{h}_{vlm}}{\|\mathbf{h}_i^{(L)}\|_2 \|\mathbf{h}_{vlm}\|_2}.
\end{equation}

Subsequently, the top-$k$ embeddings with the highest similarity scores are selected as grounding prior prompts by
\begin{equation}
\mathbf{h}_k = \text{TopK}\left(\{\mathbf{h}_i^{(L)} \mid s_i\}\right),
\end{equation}
where $\text{TopK($\cdot$)}$ selects the embeddings with the highest similarity scores.

\subsubsection{Global-Local Attention Module}\vspace{5pt}

Although the introduction of knowledge-priors enhances the guidance provided by textual localization cues, our observations indicate that textual information alone often offers limited spatial details. To compensate for the deficiency in visual awareness during the fine-tuning stages, we present a global-local attention module to strengthen the model’s focus on object regions.

\textbf{Local Attention}. Specifically, given the latent token embedding $\mathbf{h}_{box}$, a local visual cue is predicted as follows:
\begin{equation}
\mathbf{b} = f_\Phi(\mathbf{h}_{box}),
\end{equation}
where ${f_\Phi }( \cdot )$ denotes the MLP layer used for grounding box prediction. Given the local visual cue, the prompt encoder maps the geometric prompt into a sparse embedding space following the standard SAM prompt encoding scheme:
\begin{equation}
\mathbf{s}
= \operatorname{Concat}\!\Big(
\gamma\!\big(\Pi(\mathbf{b}_{tl})\big) + \mathbf{e}^{tl},\;
\gamma\!\big(\Pi(\mathbf{b}_{br})\big) + \mathbf{e}^{br}
\Big),
\end{equation}
where $\mathbf{b}_{tl}$ and $\mathbf{b}_{br}$ denote the top-left and bottom-right corner coordinates of the local visual cue, $\Pi(\cdot)$ represents coordinate normalization, $\gamma(\cdot)$ denotes the positional encoding function of the prompt encoder, $\mathbf{e}^{tl}$ and $\mathbf{e}^{br}$ are the corresponding corner-type embeddings. All parameters of the prompt encoder are kept frozen.

Accordingly, the box decoder incorporates local visual cues as attention prompts to generate the grounding representation, formulated as:
\begin{equation}
    z_{local} = \mathcal{F}_{box}(z_{enc}, \mathbf{s}).
\end{equation}

\textbf{Global Attention}. While local visual cues effectively suppress distracting regions and capture discriminative features, their emphasis on locality can weaken global context that is important for generation. To overcome this limitation, we introduce a global–local attention that jointly models complementary discriminative and generative cues, enabling more comprehensive image representations. Benefiting from its pretrained priors, the SAM decoder is able to generate a coarse key-region mask. To prevent the pretrained segmentation capability of the decoder from being degraded during fine tuning, we adopt a dual-decoder architecture. Specifically, a frozen SAM decoder is employed to generate a global mask, which is then passed through the prompt encoder to obtain dense embeddings $\mathbf{d}$. This process can be formulated as follows:
\begin{equation}
\left\{
\begin{aligned}
\mathbf{t} &= Cat(\mathbf{h}_k, \mathbf{h}_{box}),\\
\mathbf{m} &= \mathcal{F}_{dec}(z_{enc}, \mathbf{t}),\\
\mathbf{d} &
= \mathcal{P}_{mask}\!\big(\mathcal{R}(\mathbf{m})\big),
\end{aligned}
\right.
\end{equation}

where $\mathcal{R}(\cdot)$ denotes the spatial resizing operation that aligns the predicted mask with the prompt encoder resolution, $\mathcal{P}_{mask}(\cdot)$ represents the mask-to-embedding projection of the frozen prompt encoder, and $Cat$ represents concatenation. The dense embeddings are directly applied to the visual features $z_{enc}$. They are then jointly fed, together with the textual tokens, into the box decoder to produce the box prediction probability under global attention prompts. The above process can be expressed as follows:
\begin{equation}
\left\{
\begin{aligned}
z'_{enc} &= z_{enc} + \mathbf{d},\\
z_{global} &= \mathcal{F}_{box}(z'_{enc}, \mathbf{t}).
\end{aligned}
\right.
\end{equation}

Finally, we combine the global and local representation to obtain the final predicted localization bounding box, which can be formulated as follows:
\begin{equation}
    \left\{ \begin{array}{l}
{z_{box}} = \alpha  \cdot {z_{local}} + (1 - \alpha ) \cdot {z_{global}},\\
{y_b} = {f_\Phi }({z_{box}}),
\end{array} \right.
\end{equation}
where $\alpha$ is a learnable dynamic parameter that adaptively balances the relative importance of global and local information during training.

\subsection{Loss Function}

The proposed model is trained in an end-to-end manner and consists of two components: phrase identification and phrase grounding. Phrase identification is formulated as a classification task and adopt the cross-entropy loss to supervise the prediction. Specifically, given a ground-truth medical phrase $\hat y_p$, the training objective is defined as:
\begin{equation}
    {\mathcal{L}_{txt}} = {\mathcal{L}_{CE}}({\hat y_p},{y_p}).
\end{equation}

Moreover, we supervise the bounding-box prediction using a combination of two loss functions: the Smooth $\ell_{1}$ loss and the Generalized IoU (GIoU) loss. These losses jointly guide the optimization of bounding-box regression. Given a ground-truth bounding box ${\hat y_b}$, the grounding loss is formulated as:
\begin{equation}
    {\mathcal{L}_{box}} = {\mathcal{L}_{\ell_{1}}}({\hat y_b},{y_b}) + {\mathcal{L}_{giou}}({\hat y_b},{y_b}).
\end{equation}

Finally, the overall training objective is defined as the sum of the phrase identification loss and the phrase grounding loss:
\begin{equation}
    {\mathcal{L}_{total}} = {\mathcal{L}_{txt}} + {\mathcal{L}_{box}}.
\end{equation}

\begin{table*}
\centering
\renewcommand{\arraystretch}{1.3}
\caption{Performance comparison on the MRG-MS-CXR and MRG-CHESTX-RAY8 datasets.}\vspace{-0.05cm}
\label{tab:mrg_compare}
\begin{tabularx}{\textwidth}{r *{8}{>{\centering\arraybackslash}X}}
\toprule
\multirow{2}{*}{\textbf{Method}} 
& \multicolumn{4}{c}{\textbf{MRG-MS-CXR}} 
& \multicolumn{4}{c}{\textbf{MRG-CHESTX-RAY8}} \\
\cmidrule(lr){2-5} \cmidrule(lr){6-9}
& AP10 & AP30 & AP50 & mIoU 
& AP10 & AP30 & AP50 & mIoU \\
\midrule
MedRPG \cite{chen2023medical}       & 68.86 & 57.49 & 46.71 & 41.44 & 65.66 & 48.48 & 32.83 & 34.17 \\
TransVG \cite{deng2021transvg}      & 71.86 & 58.68 & 46.71 & 42.60 & 63.13 & 45.96 & 31.31 & 32.48 \\
RefTR \cite{li2021referring}        & 69.46 & 53.29 & 43.11 & 38.19 & 61.62 & 45.45 & 30.81 & 30.98 \\
VGTR \cite{9859880}         & 72.46 & 51.50 & 41.92 & 38.06 & 63.13 & 46.97 & 29.29 & 29.90 \\
LLaVA \cite{liu2023visual}        & 61.08 & 29.34 &  4.79 & 19.36 & 46.97 & 24.24 & 11.62 & 17.63 \\
InternVL \cite{chen2024internvl}      & 57.49 & 36.53 & 17.96 & 23.21 & 70.71 & 44.44 & 20.71 & 27.93 \\
LISA \cite{lai2024lisa}          
    & \uline{90.42} & \uline{77.25} & \uline{53.89} & \uline{49.17} & \uline{86.36} & \uline{59.60} & 32.32 & 38.24 \\
MedGround \cite{zou2025uncertainty}    & 87.43 & 71.26 & 51.50 & 46.04 & 82.32 & 55.05 & 31.82 & 35.91 \\
uMedGround \cite{zou2025uncertainty}   & 89.82 & 72.46 & 53.29 & 47.65 & 84.34 & 56.57 & \textbf{38.38} & \uline{38.49} \\
\textbf{KnowMRG} 
              & \textbf{91.02} & \textbf{77.25} & \textbf{56.29} & \textbf{50.31} 
              & \textbf{86.87} & \textbf{65.66} & \uline{37.88} & \textbf{41.13} \\
\bottomrule
\end{tabularx}
\end{table*}

\section{Experiments and Results}
\subsection{Experimental Settings}
\vspace{5pt}
\subsubsection{Datasets}\vspace{5pt}

Quantitative evaluation is essential for assessing grounding performance in MVG. We evaluate our method on four benchmark datasets \cite{zou2025uncertainty}, covering phrase-level grounding and downstream clinical applications.

\textbf{MRG-MS-CXR.}  
It is built upon the MIMIC-CXR dataset \cite{johnson2019mimic} and contains 1,153 triplets, each consisting of a chest X-ray, a radiology report, a diagnostic phrase, and its corresponding bounding box. The dataset covers diverse thoracic pathologies and includes 835 paired samples in total. Following \cite{zou2025uncertainty}, the data are split into training, validation, and test sets with a 7:1:2 ratio.

\textbf{MRG-ChestX-ray8.}  
This dataset is constructed from ChestX-ray8 \cite{wang2017chestx} and comprises 984 samples. The corresponding radiology reports are generated using GPT-4. We adopt the same 7:1:2 split setting for training, validation, and testing.

\textbf{MRG-MIMIC-VQA.}  
To evaluate grounding in a clinical question-answering scenario, VQA instances aligned with MRG-MS-CXR are extracted from the Medical-Diff-VQA \cite{hu2023expert}. The resulting subset contains 158 paired VQA samples.

\textbf{MRG-MIMIC-Class.}  
It targets category-level localization and is constructed by selecting disease categories relevant to MRG-MS-CXR from MIMIC-CXR. Category descriptions are generated by GPT-4 as textual prompts, yielding 163 category-level localization instances corresponding to the MRG-MS-CXR test set after filtering non-overlapping samples.

\subsubsection{Evaluation Metrics}\vspace{5pt}

To evaluate the effectiveness of MRG, we adopt evaluation metrics to assess the accuracy of bounding-box grounding. We employ the widely used mean Intersection over Union (mIoU) metric for comprehensive comparison. Following the protocol of uMedGround \cite{zou2025uncertainty}, we further report the average precision (AP) for bounding-box prediction. Specifically, a predicted bounding box is considered a true positive if its mIoU with the ground-truth box exceeds a predefined threshold (0.1, 0.3, or 0.5). Accordingly, we denote these thresholds as AP10 (mIoU$>$0.1), AP30 (mIoU$>$0.3), and AP50 (mIoU$>$0.5), respectively.

\subsubsection{Baselines}\vspace{5pt}

We compare our proposed method with several general-purpose and medical-specific visual grounding approaches, including RefTR \cite{li2021referring}, VGTR \cite{9859880}, TransVG \cite{deng2021transvg}, and MedRPG \cite{chen2023medical}. 
These baselines require phrase extraction from medical reports during inference, which is performed using GPT-4 \cite{nori2023capabilities}.
In addition, we compare against widely used VLMs, including gpt4o \cite{achiam2023gpt}, LLaVA-13B \cite{liu2023visual}, and InternVL2-8B \cite{chen2024internvl,chen2024far}. Notably, they are further fine-tuned on task-specific datasets using LoRA adapters. All baseline methods adopt their official configurations and ensure that training and evaluation are conducted under consistent experimental conditions.

\begin{figure*}
	\centering
	\includegraphics[width=0.99\textwidth]{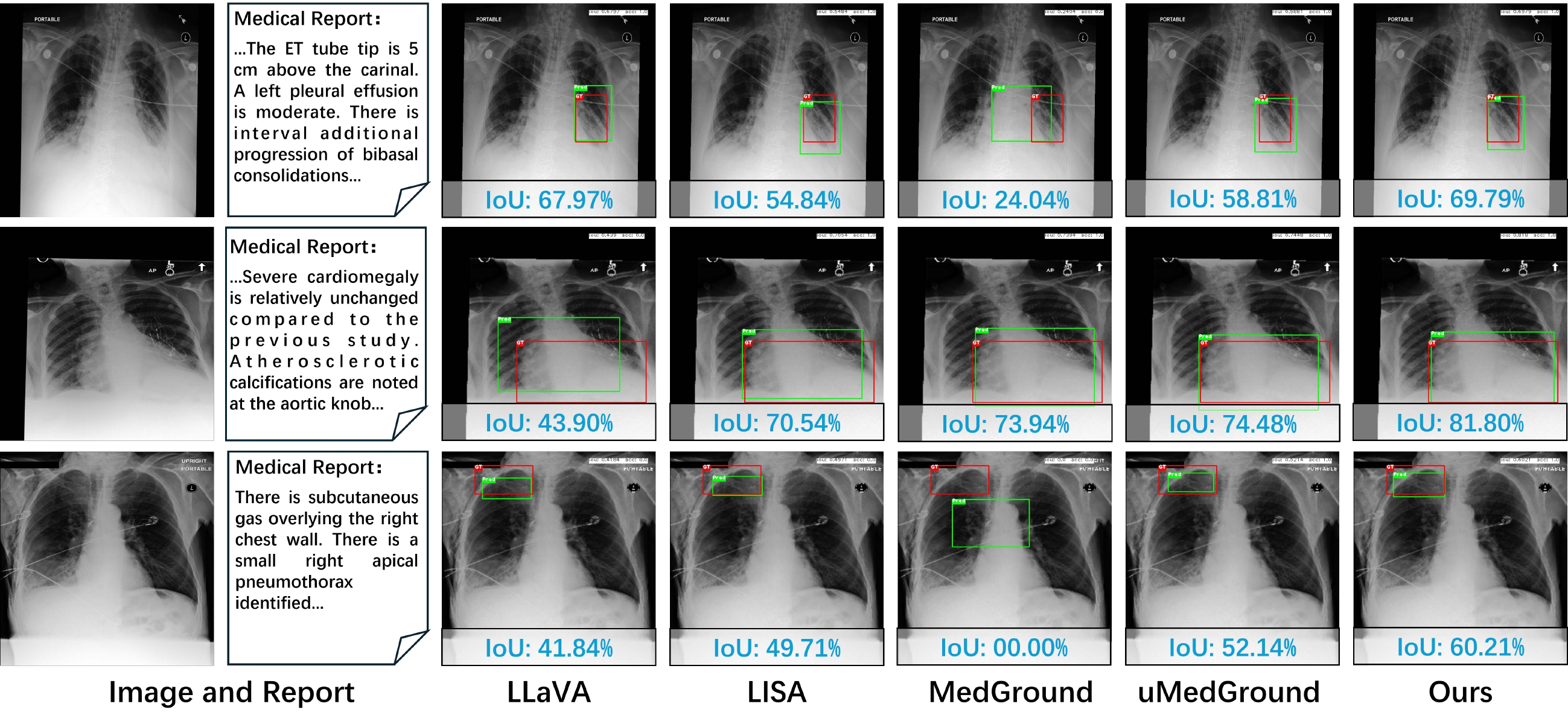} 
	\caption{Visual comparison of medical report grounding results across different methods on the MRG-MS-CXR dataset.}
	\label{compare1}
\end{figure*}

\subsubsection{Implementation Details}\vspace{5pt}

All experiments are conducted using two NVIDIA RTX 4090 GPUs with 24 GB memory. The training pipeline is implemented based on the DeepSpeed engine \cite{kirillov2023segment}. We employ the AdamW optimizer \cite{bannur2024maira}, with the learning rate and weight decay set to 0.0003 and 0, respectively. A WarmupDecayLR scheduler is adopted, with the number of warm-up iterations set to 100. The per-device batch size is set to 2. During training, we report results based on the best-performing checkpoints on the validation set, selected under different random seeds for both baseline methods and our proposed model.

\begin{figure*}
	\centering
	\includegraphics[width=0.99\textwidth]{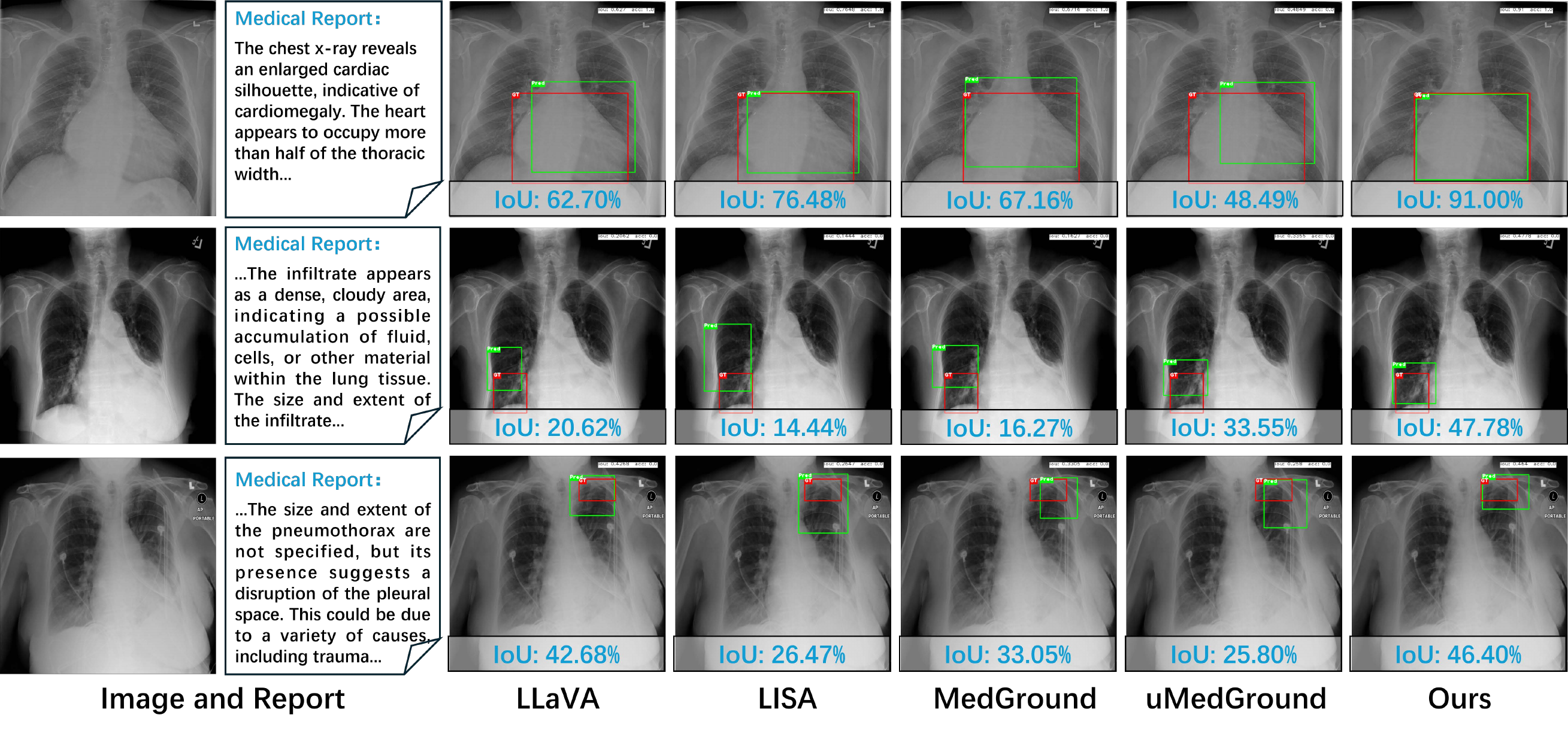} 
    \vspace{-0.05cm}
	\caption{Visual comparison of medical report grounding results across different methods on the MRG-ChestX-ray8 dataset.}
    \vspace{-0.0cm}
	\label{compare2}
\end{figure*}

\subsection{Experimental Results on MRG-MS-CXR Dataset}
\vspace{5pt}
\subsubsection{Phrase Grounding}
\vspace{5pt}

Table~\ref{tab:mrg_compare} summarizes phrase grounding performance on the MRG-MS-CXR dataset. Among traditional phrase grounding methods, TransVG consistently outperforms RefTR and VGTR, while MedRPG achieves comparable results. However, all these approaches exhibit limited performance on AP50 and mIoU, with mIoU remaining below $45\%$, indicating insufficient spatial precision. This limitation mainly stems from the coarse semantic granularity and weak spatial specificity of diagnostic phrases extracted directly from raw medical reports. Compared to traditional methods, VLMs based methods achieves AP30 above $70\%$ and mIoU exceeding $45\%$. uMedGround introduces uncertainty-aware prediction to enhance stability and performance. Similarly, LISA adopts an end-to-end paradigm and benefits from rich language priors during decoder pretraining, yielding the strongest performance among baseline methods.
Despite these advances, our method achieves the best overall results on MRG-MS-CXR. By introducing tailored prompt modules, it attains an AP10 above $91\%$, improves AP50 to $56.29\%$, and reaches an mIoU of $50.31\%$. Notably, our model consistently outperforms state-of-the-art methods under higher overlap thresholds, demonstrating superior discriminative capability for precise medical report–driven grounding.

\subsubsection{Visual Comparisons}\vspace{5pt}

Fig.~\ref{compare1} shows grounding results for three representative cases, including left perihilar opacity, enlarged cardiac silhouette, and right apical pneumothorax. Using complete medical reports as input, we compare our method with representative baselines to highlight differences in localization accuracy. Across the three cases, baseline methods without explicit anatomical priors are often distracted by globally salient regions, leading to shifted, missed, or overly loose bounding boxes, especially in complex backgrounds or when lesion boundaries are ambiguous. In contrast, our method leverages anatomy-related knowledge priors to constrain the global search space and further refines critical regions through local prompts, resulting in more compact, spatially coherent, and ground-truth-aligned localization across all cases. 

\begin{table*}[htbp]
\centering
\caption{Performance comparison on the MRG-MIMIC-VQA and MRG-MIMIC-Class datasets. 
}
\label{tab:mrg_vqa_class}
\begin{tabularx}{\textwidth}{r *{8}{>{\centering\arraybackslash}X}}
\toprule
\multirow{2}{*}{\textbf{Method}} 
& \multicolumn{4}{c}{\textbf{MRG-MIMIC-VQA}} 
& \multicolumn{4}{c}{\textbf{MRG-MIMIC-CLASS}} \\
\cmidrule(lr){2-5} \cmidrule(lr){6-9}
& AP10  & AP30  & AP50  & mIoU 
& AP10 & AP30  & AP50  & mIoU  \\
\midrule
LLaVA \cite{liu2023visual}
& 83.54 & 63.92 & 39.24 & 40.28
& 84.81 & 65.51 & 40.19 & 40.35 \\

Lisa \cite{lai2024lisa}
& \uline{87.97} & \uline{75.32} & \uline{55.06} & \uline{47.90} 
& \uline{90.19} & \uline{77.22} & \uline{55.06} & \uline{49.00} \\

MedGround \cite{zou2025uncertainty}
& 87.34 & 68.99 & 49.37 & 44.18 
& 85.44 & 70.25 & 51.27 & 44.63 \\

uMedGround \cite{zou2025uncertainty}
& 87.97 & 73.42 & 49.37 & 46.61 
& 88.47 & 73.39 & 50.55 & 47.39 \\

\textbf{KnowMRG}
& \textbf{90.51} & \textbf{77.22} & \textbf{55.06} & \textbf{49.61}
& \textbf{91.77} & \textbf{77.53} & \textbf{56.65} & \textbf{50.40} \\
\bottomrule
\end{tabularx}
\end{table*}

\begin{figure*}[t]
	\centering
	\includegraphics[width=0.99\textwidth]{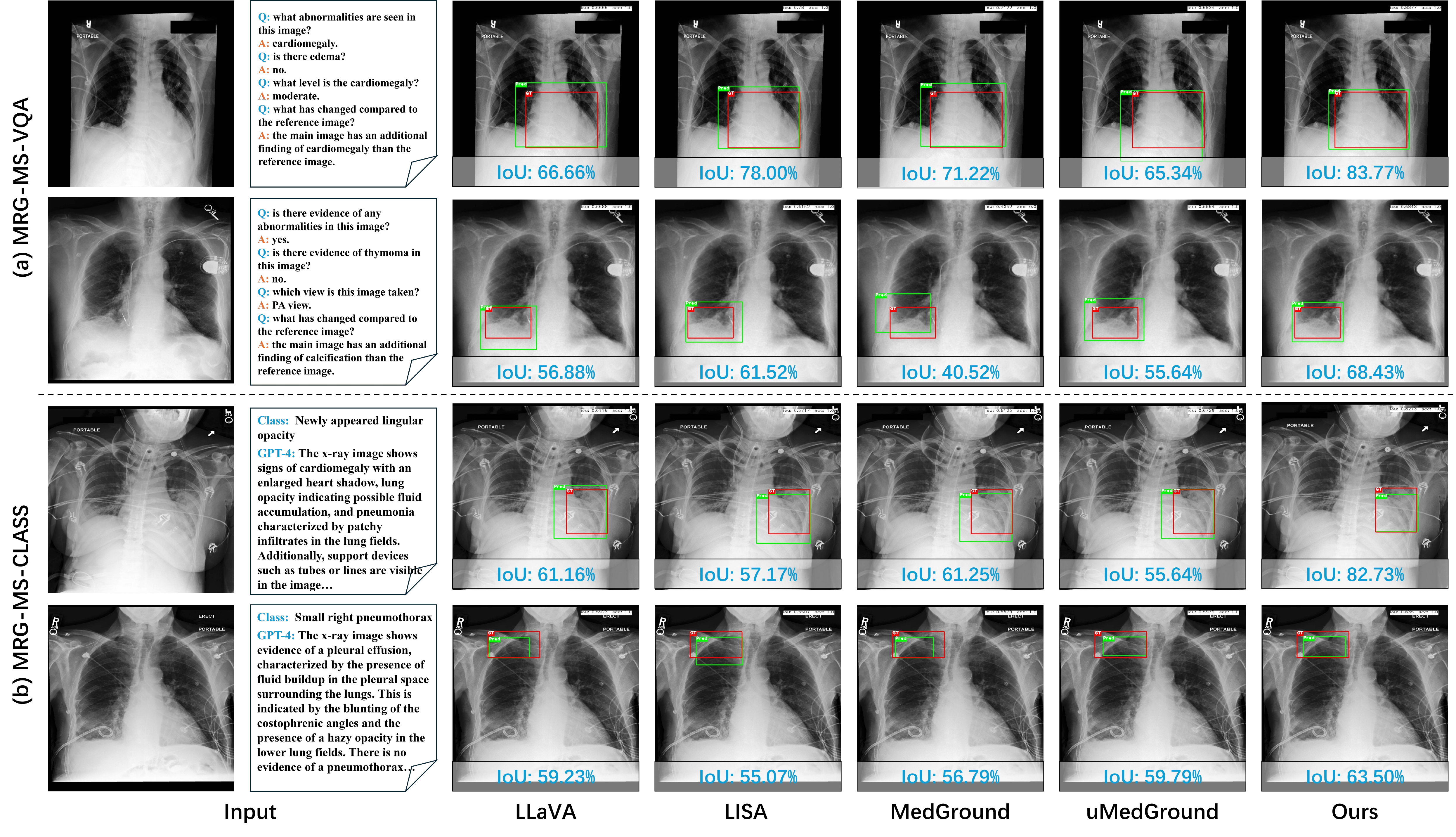} 
	\caption{Visual comparison results across different methods under VQA-based and category-level clinical text inputs.}
    \vspace{-0.0cm}
	\label{clinical}
\end{figure*}

\subsection{Experimental Results on MRG-ChestX-ray8 Dataset}
\vspace{5pt}
\subsubsection{Phrase Grounding}
\vspace{5pt}

Table~\ref{tab:mrg_compare} reports phrase grounding results on the MRG-ChestX-ray8 dataset. Traditional phrase grounding methods exhibit a pronounced performance degradation under this setting. Although MedRPG maintains relatively higher AP10 and AP30 scores, its AP50 and mIoU drop to $32.83\%$ and $34.17\%$, respectively, indicating limited robustness when report semantics are imprecise. TransVG, RefTR, and VGTR follow similar trends, with performance further deteriorating at medium and high IoU thresholds, highlighting the strong dependence of phrase-based grounding paradigms on text quality. 
General VLMs like LLaVA and InternVL perform reasonably at lenient thresholds but drop sharply under stricter localization, reflecting a lack of precise spatial grounding without radiology supervision. uMedGround mitigates GPT-induced uncertainty, and LISA maintains stable performance, demonstrating the value of joint multimodal-visual modeling. In constrast, our method achieves the best overall performance on MRG-ChestX-ray8. Specifically, it reaches AP10 and AP30 scores of 86.87\% and 65.66\%, respectively, while further improving mIoU to 41.13\%. These results indicate that, by explicitly guiding model attention toward clinically relevant regions through carefully designed prompting strategies, our approach remains robust even under highly noisy settings where textual supervision relies entirely on generated reports.

\subsubsection{Visual Comparisons}\vspace{5pt}

Fig.~\ref{compare2} presents grounding results for three representative cases from the MRG-ChestX-ray8 dataset, including cardiomegaly, infiltrate, and pneumothorax. 
We compare LLaVA, LISA, MedGround, uMedGround, and the proposed method under identical input reports.
It can be observed that the comparison methods often exhibit imprecise localization, particularly when abnormalities are small. While some approaches can roughly identify relevant areas, they frequently suffer from localization shifts. In contrast, our method consistently generates boxes that are well aligned with the annotated regions, achieving more complete abnormality coverage while effectively suppressing irrelevant background responses. As a result, it produces more compact and accurate localization across diverse abnormality types and difficulty levels. 
Overall, the qualitative results are consistent with the quantitative findings in Table~\ref{tab:mrg_compare}, demonstrating that the proposed method delivers more accurate and clinically meaningful grounding.

\subsection{Clinical Extension}\vspace{5pt}

To evaluate generalization and clinical applicability, we extend our experiments to two additional settings: medical visual question answering and category-level localization. MRG-MIMIC-VQA uses medical question–answer pairs as textual inputs, reflecting real-world clinical consultations, whereas MRG-MIMIC-CLASS provides only disease category descriptions, representing weakly specified clinical scenarios without detailed reports.

\subsubsection{Qualitative Comparison}\vspace{5pt}

\begin{figure*}[!t]
	\centering
	\includegraphics[width=0.99\textwidth]{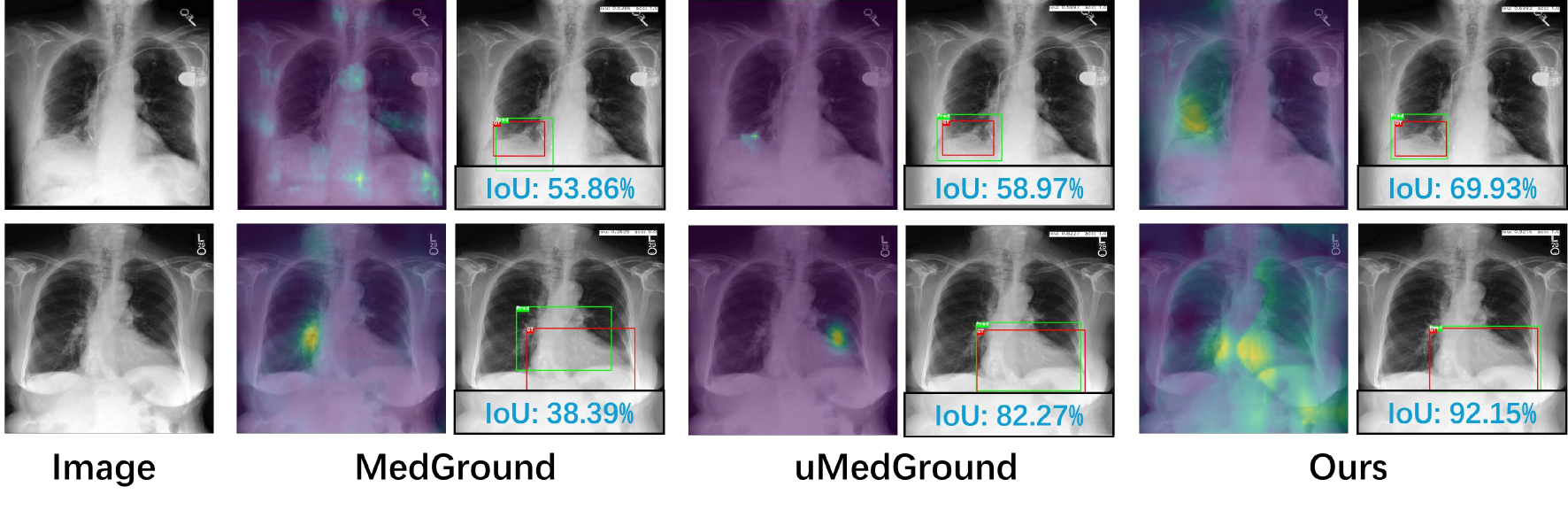} 
    \vspace{-0.05cm}
	\caption{Visualization of attention maps obtained by different grounding methods.}
	\label{attn}
\end{figure*}

Table~\ref{tab:mrg_vqa_class} summarizes the quantitative results. On MRG-MIMIC-VQA, our method achieves the best performance across all metrics, clearly outperforming LISA and uMedGround. While existing methods show reasonable gains under lower overlap thresholds, their performance drops markedly on stricter metrics such as AP50 and mIoU, revealing limited boundary localization accuracy. A similar trend is observed on MRG-MIMIC-CLASS. Despite relying only on category-level textual inputs, our method consistently outperforms competing approaches and attains an mIoU of 50.40\%. This indicates that KnowMVG effectively captures intrinsic associations between category semantics and visual evidence, enabling stable and accurate localization under weak textual supervision.

Further analysis shows that variations in text formulation significantly affect the performance of LISA and uMedGround, suggesting a strong dependence on textual prior rather than visual semantics. These results demonstrate the adaptability and robustness of our method across diverse clinical input forms, highlighting its potential for practical deployment in real-world clinical scenarios.

\subsubsection{Visual Comparison}\vspace{5pt}

Fig.~\ref{clinical} presents visual comparisons on the MRG-MIMIC-VQA and MRG-MIMIC-CLASS tasks. In the MRG-MIMIC-VQA setting, baseline methods can roughly identify abnormal regions but often produce loosely bounded boxes with inaccurate spatial alignment. Although MedGround and uMedGround improve localization to some extent, noticeable boundary deviations remain, especially for cardiomegaly and subtle thoracic abnormalities. In contrast, our method consistently generates compact bounding boxes that closely align with the ground truth, achieving the highest IoU across representative cases and demonstrating effective translation from VQA-derived textual cues to precise spatial localization. 

In the MRG-MIMIC-CLASS setting, where only category-level disease descriptions are provided, baseline localization degrades significantly, with predicted regions often deviating from actual lesions. In contrast, our approach maintains robust, well-aligned predictions despite the absence of explicit localization cues, demonstrating its ability to associate high-level disease semantics with discriminative visual patterns.

\begin{table}[!t]
\centering
\renewcommand{\arraystretch}{1.3}
\caption{Ablation study of key components in our model on the MRG-MS-CXR and MRG-CHESTX-RAY8 datasets.}
\label{tab:ablation_kg_glb}
\resizebox{\linewidth}{!}{
\begin{tabular}{cc cccccccc}
\toprule
\multirow{2}{*}{\textbf{KPS}} 
& \multirow{2}{*}{\textbf{GLA}}
& \multicolumn{4}{c}{\textbf{MRG-MS-CXR}} 
& \multicolumn{4}{c}{\textbf{MRG-CHESTX-RAY8}} \\
\cmidrule(lr){3-6} \cmidrule(lr){7-10}
& 
& AP10  & AP30  & AP50  & mIoU 
& AP10  & AP30  & AP50  & mIoU  \\
\midrule
&
& 87.43 & 71.26 & 51.5 & 46.04 
& 82.32 & 55.05 & 31.82 & 35.91 \\

\checkmark & 
& \textbf{93.41} & 74.25 & 56.29 & 49.67
& 85.35 & 62.12 & 36.36 & 39.61 \\

& \checkmark
& 91.02 & 74.25 & 56.29 & 49.57
& \textbf{88.38} & 59.60 & \textbf{37.88} & 40.15 \\

\checkmark & \checkmark
& 91.02 & \textbf{77.25} & \textbf{56.29} & \textbf{50.31}
& 86.87 & \textbf{65.66} & \textbf{37.88} & \textbf{41.13} \\
\bottomrule
\end{tabular}
}
\end{table}

\begin{table}[!t]
\centering
\renewcommand{\arraystretch}{1.3}
\caption{Ablation study of key components in our model on the MRG-MIMIC-VQA and MRG-MIMIC-Class datasets.}
\label{tab:ablation_kg_glb_vqa_class}
\resizebox{\linewidth}{!}{
\begin{tabular}{cc cccccccc}
\toprule
\multirow{2}{*}{\textbf{KPS}} 
& \multirow{2}{*}{\textbf{GLA}}
& \multicolumn{4}{c}{\textbf{MRG-MIMIC-VQA}} 
& \multicolumn{4}{c}{\textbf{MRG-MIMIC-CLASS}} \\
\cmidrule(lr){3-6} \cmidrule(lr){7-10}
& 
& AP10  & AP30  & AP50  & mIoU 
& AP10  & AP30  & AP50  & mIoU  \\
\midrule
&
& 87.34 & 68.99 & 49.37 & 44.18 
& 85.44 & 70.25 & 51.27 & 44.63 \\

\checkmark & 
& 89.24 & 73.73 & 54.11 & 48.57
& 87.34 & 72.15 & 53.80 & 46.90 \\

& \checkmark
& \textbf{91.46} & 75.00 & 56.01 & 49.58
& 87.97 & 75.32 & 55.06 & 49.33 \\

\checkmark & \checkmark
& 87.97 & \textbf{75.95} & \textbf{56.33} & \textbf{49.93}
& \textbf{91.77} & \textbf{77.53} & \textbf{56.65} & \textbf{50.40} \\
\bottomrule
\end{tabular}
}
\end{table}

\subsection{Ablation Study and Analysis}\vspace{5pt}

In this section, we conduct an ablation study to investigate the impact of the proposed Knowledge-enhanced Prompting Strategy (KPS) and Global-Local Attention (GLA) module. The corresponding results are summarized in Table~\ref{tab:ablation_kg_glb_vqa_class} and Table~\ref{tab:ablation_kg_glb}.

\subsubsection{Effectiveness of Knowledge-enhanced Prompting Strategy}\vspace{5pt}

We first consider a baseline configuration that excludes all proposed components. As shown in Table~\ref{tab:ablation_kg_glb_vqa_class}, this setting yields limited performance across all four datasets, particularly on stringent metrics such as AP50 and mIoU. These results indicate that without explicit medical knowledge modeling and semantic guidance, the model struggles to achieve precise localization of abnormal regions. We then incorporate the KPS to enhance the modeling of medical concepts and their relationships. As shown in Table~\ref{tab:ablation_kg_glb}, enabling KPS alone leads to consistent performance improvements across all datasets, with more pronounced gains observed on AP10 and AP30. This suggests that structured medical knowledge effectively complements textual semantics by providing discriminative high-level priors. However, the improvements on AP50 and mIoU remain relatively modest, indicating that knowledge enhancement alone is insufficient for fine-grained boundary alignment.

\subsubsection{Effectiveness of Global–Local Attention Module}\vspace{5pt}

Building upon the KG-enhanced setting, we further introduce the GLA, which explicitly models the interaction between global semantic representations and local visual cues. When KPS and GLA are jointly enabled, the model achieves the best performance across all four benchmarks, with particularly notable improvements on AP50 and mIoU. These results demonstrate that the GLA module effectively compensates for the limitations of knowledge-only enhancement by facilitating fine-grained spatial alignment. Specifically, while the KPS strengthens semantic understanding of medical concepts, the GLA module bridges semantic representations and localized visual perception. The two components are therefore complementary and work synergistically, enabling enhanced robustness and consistency across diverse text input forms and clinical task settings.

\subsubsection{Attention Analysis}\vspace{5pt}

To further investigate the grounding behavior of different methods, we provide an attention-based comparison as shown in Fig.~\ref{attn}. The visualization highlights the attention distributions of MedGround, uMedGround, and the proposed method on representative cases. MedGround produces relatively diffuse attention maps that are weakly concentrated around the pathological regions, resulting in suboptimal overlap with the ground-truth boxes. Although uMedGround improves spatial concentration in certain cases, its attention responses remain partially scattered or biased toward surrounding anatomical structures, leading to inconsistent grounding performance. In contrast, the proposed method generates more focused and well-aligned attention patterns that closely correspond to the true areas. The highlighted regions are spatially compact and semantically consistent with the abnormal findings. This observation suggests that the proposed strategy effectively guides the model to attend to clinically relevant regions while suppressing irrelevant background responses.

\section{Conclusion}

In this work, we investigate the problem of medical visual grounding and analyze the limitations of existing VLM–based approaches in precise visual grounding. We observe that relying solely on latent token embeddings provides insufficient localization attention, leading to inaccurate or unstable grounding results in complex clinical scenarios. To address this issue, we propose a novel framework that introduces a knowledge-enhanced prompting strategy and global–local attention module to enhance spatial awareness during decoding. The knowledge prior encodes phrase-related localization information into compact embeddings, while the global–local attention jointly leverages global visual representation and local visual cues to guide the model toward clinically relevant regions. Extensive experiments on multiple MVG benchmarks demonstrate that the proposed method achieves consistent performance improvements over existing approaches across different text input settings, indicating its robustness and potential applicability in diverse clinical scenarios.



\ifCLASSOPTIONcaptionsoff
  \newpage
\fi

\bibliographystyle{IEEEtran}
\bibliography{TMI_bib_v2}

\end{document}